\documentclass[runningheads]{llncs}
\usepackage{lipsum}
\usepackage{graphicx}
\usepackage[dvipsnames]{xcolor}
\begin{document}
\title{A Comparative Analysis of Feature Selection Methods for Biomarker Discovery in Study of Toxicant-treated Atlantic Cod (\emph{Gadus morhua}) Liver}
\titlerunning{A Comparative Analysis of Feature Selection Methods}
%
\author{Xiaokang Zhang\orcidID{0000-0003-4684-317X} \and
Inge Jonassen\orcidID{0000-0003-4110-0748}}
\authorrunning{X. Zhang et al.}
%
\institute{Computational Biology Unit, Department of Informatics, University of Bergen
\email{\{xiaokang.zhang, inge.jonassen\}@uib.no}\\
\url{https://www.cbu.uib.no/jonassen/}}

\maketitle              
\begin{abstract}
Univariate and multivariate feature selection methods can be used for biomarker discovery in analysis of toxicant exposure. Among the univariate methods, differential expression analysis (DEA) is often applied for its simplicity and interpretability. A characteristic of methods for DEA is that they treat genes individually, disregarding the correlation that exists between them. On the other hand, some multivariate feature selection methods are proposed for biomarker discovery. Provided with various biomarker discovery methods, how to choose the most suitable method for a specific dataset becomes a problem. In this paper, we present a framework for comparison of potential biomarker discovery methods: three methods that stem from different theories are compared by how stable they are and how well they can improve the classification accuracy. The three methods we have considered are: Significance Analysis of Microarrays (SAM) which identifies the differentially expressed genes; minimum Redundancy Maximum Relevance (mRMR) based on information theory; and Characteristic Direction (GeoDE) inspired by a graphical perspective. Tested on the gene expression data from two experiments exposing the cod fish to two different toxicants (MeHg and PCB 153), different methods stand out in different cases, so a decision upon the most suitable method should be made based on the dataset under study and the research interest.

\end{abstract}
\section{Introduction}

Atlantic cod (\emph{Gadus morhua}) is one of the most important commercial fish species in Norway~\cite{Ageeva17}, forming the basis for fisheries, trade, and, historically, civilization. Unfortunately, cod is increasingly susceptible to marine pollution from petroleum activities~\cite{Goksoyr91,Balk11}. Atlantic cod is commonly used as an indicator species in marine environmental monitoring programs, and a useful model organism to investigate the effect of toxicants~\cite{Sundt12,Chesman07,Olsvik09}. Finding the best set of biomarkers for Atlantic cod exposed to toxicants is of high research and commercial value. Biomarkers can for example be defined based on the expression level of a set of genes or proteins. Biomarker discovery is an essential part in study of toxicant exposure, and many methods have been proposed to find biomarkers~\cite{Robotti14}. However, a remaining question is, provided with numbers of biomarker discovery methods, which method is the most suitable one for a particular dataset. This paper provides a framework to compare potential biomarker discovery methods and to give researchers a better basis for choosing which one to use for the task at hand. \par

In the context of statistics and machine learning, biomarker discovery corresponds to a feature selection problem, where the purpose is to identify the most distinguishing features, for example, distinguishing normal and toxicant-treated cod livers. The task of feature selection is to identify, from a wide range of features, those that are best suited for classification.\par

The strategies of feature selection methods can be divided into two categories~\cite{Robotti14}:

\begin{enumerate}
\item Classical univariate statistical methods, where the features are considered as independent from each other. Genes that are differentially expressed are regarded as biomarkers.
\item Multivariate methods, which take the interaction between features into consideration when selecting the important features allowing to distinguish samples coming from different groups.
\end{enumerate}

The classical univariate methods try to find the features having significantly different values between the different groups, e.g. control group and treated group. One of the most popular and basic methods is Student's t-test~\cite{De Winter13}. Some similar research also adopted Analysis of Variance (ANOVA) and Significance Analysis of Microarrays (SAM) to find the differential expressed genes~\cite{Tusher01,Yadetie13,Yadetie14,Yadetie16,Yadetie17}. A main drawback of such approaches is that they rest on the assumption that all the genes or proteins are independent from each other, which is clearly not true, since both genes and proteins are part of a biological system where they interact with each other~\cite{Shannon03,Tong04}. \par

On the other hand, multivariate methods will take the interaction among features into consideration, reflecting that the features are acting in groups. Many feature selection and machine learning methods try to find the features most correlated with the class labels and take the interaction among features into consideration at the same time. \par

Feature selection methods are often divided into three categories: filter methods which focus on the relation between feature values and class labels; wrapper methods which use an objective function (can be the classification accuracy of the classifier) to evaluate features; and embedded methods where the classifier selects the features automatically~\cite{He10}. The latter two are both classifier-dependent, and filter methods are more like a one-way decision without feedback from prediction accuracy. In order to find a more general feature selection method, which does not only work well with one specific classifier, we will only focus on the filter methods. \par

In toxicant exposure study, or more generally, in the context of biology, very often, researchers are faced with the high-dimension-small-sample-size issue, since it is hard and expensive to get a high number of samples (it is often around 10 or even lower), but the number of features (genes or proteins) is usually very high (over one thousand). In such cases, two problems are difficult to avoid: finding a reliable feature subset, as in this case the possibility of chance correlation is quite high; assuring that the selected features are true biomarkers. The true biomarkers should be data-independent, meaning that a small change in the samples should not lead to a large change in the selected features, which requires the feature selection method to be stable. Besides of that, they should also be qualified to be treated as the representatives of the whole feature list and should therefore be able to improve a classifier's prediction accuracy while classifying samples from different biological conditions. Therefore, we will compare the feature selection methods based on two aspects of their performance: stability to find a reliable feature subset and ability to improve a classifier's prediction accuracy. \par

To make the work reproducible, all the data sets and source codes are publicly available at \url{https://github.com/zhxiaokang/FScompare}.


\section{Methods}
\subsection{Data sets}
Two datasets from study of toxicant-treated Atlantic cod liver are used here. One is from the study of the hepatic proteome of MeHg-exposed Atlantic cod, where there are 10 samples in control group, 9 samples in low-dose treated group (0.5 mg/kg Body Weight MeHg), and 9 samples in high-dose treated group (2 mg/kg BW MeHg). The abundances of 1143 proteins were measured after the samples were exposed in vivo to MeHg for two weeks~\cite{Yadetie16}. The other study is from the quantitative proteomics analysis of Atlantic cod livers treated with PCB 153 of various doses of PCB 153 (0, 0.5, 2 and 8 mg/kg BW PCB 153) for two weeks. There are 10 samples in each control group, low-dose treated group, medium-dose treated group, and high-dose treated group. Then 1272 liver proteins are quantified~\cite{Yadetie17}.\par

\subsection{Principle of method and notations}
Consider a set of $m$ samples \{$x_i, y_i$\} ($i=1, 2, ... m$). Each sample has $n$ input variables $x_{i, j}$ ($j=1, 2, ... n$) and one output variable $y_i$. From the original feature set $F$, a feature selection method will select a subset $S$ of $k$ variables. 

Suppose that there are $P$ feature selection methods to be compared. Using Leave-One-Out Cross-Validation (LOOCV), $m$ feature subsets will be generated for each pre-defined value of $k$. The stability of each feature selection method $Stab_{p, k}$ ($p=1, 2, ... P$) can be calculated based on those $m$ subsets. 

To test their ability to improve a classifier's prediction accuracy, the generated feature subsets will then be applied to train a classifier and the prediction accuracy of the corresponding classifier will also be measured. Area Under the Curve (AUC) is used to measure the classifier's prediction accuracy~\cite{Fawcett06}. If tested on $Q$ classifiers, the prediction accuracy of each classifier can be calculated $AUC_{p, q, k}$ ($q=1, 2, ... Q$). Considering both matrices $Stab$ and $AUC$, a general evaluation of each feature selection method can finally be achieved so that researchers can choose a proper method for their data.

But the stability does not necessarily agree with the prediction accuracy: the most stable feature selection method may not achieve the highest prediction accuracy. Then the researchers need to balance between these two measures according to their preference and the needs of the project.

\subsection{Feature selection methods}

Some representatives of those two strategies (univariate and multivariate) are compared. 
For the univariate methods, SAM is applied here, since it was used in the literature from where our data comes. SAM was designed to identify genes with significantly differential expression in microarray experiments.
For the multivariate methods, we utilize minimum Redundancy Maximum Relevance (mRMR)~\cite{Peng05} and Characteristic Direction from a geometrical aspect (GeoDE)~\cite{Clark14}.
mRMR is based on information theory. It tries to find out the feature subset in which the redundancy among the features are minimized and the relevance of features and the targeted classes are maximized.
GeoDE uses linear discriminant analysis to define a separating hyperplane and the orientation of the hyperplane is used to identify the differentially expressed genes.

Those methods are selected for our comparison because they are based on different theories so that our results are more likely to be valid in general, and they are all widely used biomarker discovery methods. So $P$ equals $3$ in this case, but researchers can always compare as many feature selection methods as they want. \par

\subsection{Performance measurement}

Performance of feature selection methods is measured by two factors: stability and accuracy.

Many measures of stability have been proposed. Nogueira et al. studied 15 different measures proposed between 2002 and 2018 and also proposed their novel measure~\cite{Nogueira18}. In our case where the purpose is to compare the stability of different feature selection methods, the absolute values of stability are not that important as long as they are comparable for different methods under the same settings. In each round of comparison, the number of selected features $k$ is a constant, so the stability measure does not need to be able to cope with various numbers of features. LOOCV will generate more than two feature sets based on which the stability is calculated, so the measures which are defined for a pair of feature sets are not proper choices. Considering the measures that satisfy all the requirements, we chose StabPerf~\cite{Davis06} for its simplicity and interpretability. The stability is defined as:
\begin{equation}
    Stab_{p,k}=\frac{\sum_{f\in{F}}(freq(f)/m)}{|F|}
\end{equation}
Where $Stab_{p,k}$ is the stability of a given feature selection method $p$ with a pre-defined $k$; $m$ is the number of feature subsets analyzed; $F$ is the set of features that appear in at least one of the $m$ subsets and $|F|$ indicates the cardinality of $F$; $freq(f)$ is the frequency of feature $f\in{F}$ that appears in those $m$ subsets.

To test the ability to improve a classifier's prediction accuracy, four popular classification methods are utilized here: Random Forest (RF)~\cite{Breiman01}, Support Vector Machine (SVM)~\cite{Cortes95}, and extended two-class logistic regression (RIDGE and LASSO are applied)~\cite{Friedman10}.

\subsection{Cross-validation approach}

We characterize our problem as a two-class classification problem: the control group versus the treated group. In the process of classification, we need to divide the samples into training set and testing set. But since the number of samples is quite limited, we apply the strategy of LOOCV, which means that in every training-prediction process, we leave one sample out as testing set, and use the other samples as training set to search for the most important features and to train a classifier. With $m$ samples, we will use the $i^{th}$ sample to test the prediction accuracy of the classifier trained from the other $m-1$ samples. The average of performance observed over all $m$ predictions will be regarded as the estimate of the performance of the model trained over the whole sample set. To avoid overfitting or an overly optimistic estimate, it should be noted that the feature selection and training of classifiers are only limited to the training set, to avoid the information from the testing set leaking into the model training procedure~\cite{Cawley10}. 
That makes the size of testing set decided by the number of samples in one classification problem, e.g. 19 in MeHg's high-dose case. Moreover, 19 samples indicate 19 rounds of feature selection and prediction, resulting in 19 selected feature subsets and 19*4 classifiers. Therefore, if a feature selection method is stable enough, there should be a big overlap among these 19 selected feature subsets; at best the feature subsets would be identical. And if the selected features are true biomarkers, the resulting 76 classifiers should yield high prediction accuracies. \par

To make our comparison more stable, avoiding the accidental findings, and to analyze the characteristic of the feature selection methods, we repeat the above process with different numbers of selected features (ranging from 40 to 400 with a step of 40, but also including 12 and 24 to look into more details with small numbers of selected features where the output varies a lot).\par 

Tukey's Honestly Significant Difference Test (Tukey HSD Test)~\cite{Yandell17} is also applied to test the significance of the differences between different methods' performance on stability and prediction accuracy. \par

\begin{figure*}
\center
\includegraphics[width=0.65\textwidth]{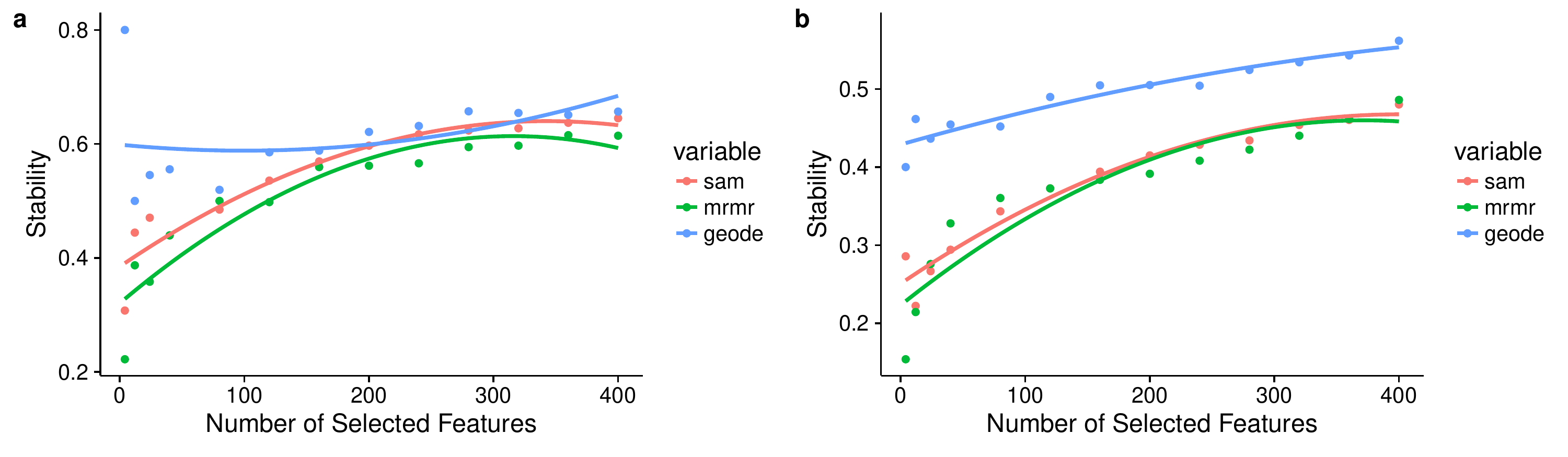}
\caption{Stability of feature selection methods on MeHg data.
      (a) Experiment on high-dose group versus control group.
      (b) Experiment on low-dose group versus control group.}\label{fig:01}
\end{figure*}

\begin{figure*}
\center
\includegraphics[width=0.995\textwidth]{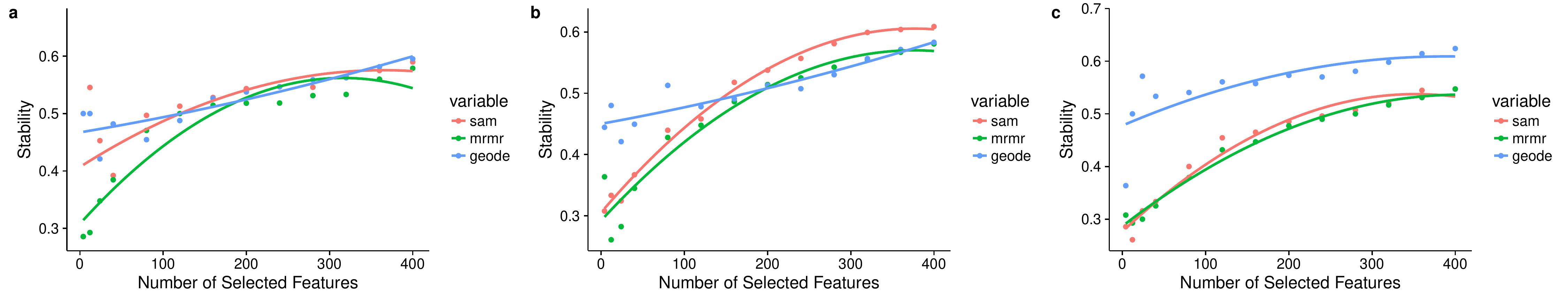}
\caption{Stability of feature selection methods on PCB 153 data.
      (a) Experiment on high-dose group versus control group.
      (b) Experiment on medium-dose group versus control group.
     (c) Experiment on low-dose group versus control group.}\label{fig:02}
\end{figure*}



\section{Results}

\subsection{Stability}

We can see from Fig.~1\vphantom{\ref{fig:01}} and Fig.~2\vphantom{\ref{fig:02}} that the performance of GeoDE is more stable than SAM and mRMR across different numbers of selected features (with the smallest variance). Another big difference between GeoDE and the other two methods can be seen in low-dose condition of both MeHg and PCB 153: with all numbers of selected features, GeoDE consistently outperforms SAM and mRMR (Fig.~1\vphantom{\ref{fig:01}}b, Fig.~2\vphantom{\ref{fig:02}}c). \par

The results from Tukey HSD Test on stability are shown in Table~\ref{Tab:01}. We limit the family error rate to 0.05, so the cases with an adjusted p-value (p-adj) smaller than 0.05 are regarded as significantly different. In accordance with the previous analysis, in low-dose condition both for MeHg and PCB 153, GeoDE is much more stable than the other two feature selection methods. \par

\begin{table}
\centering
\caption{Tukey HSD Test on stability}\label{Tab:01}
\begin{tabular}{@{}cccccc} \hline
Toxicant & Dose condition & \multicolumn{3}{c}{Comparison} & p-adj\\ \hline
MeHg & low & GeoDE & is better than & SAM & 0.0006\\
MeHg & low & GeoDE &  is better than & mRMR & 0.0005\\
PCB 153 & low & GeoDE & is better than & SAM & 0.0014\\
PCB 153 & low & GeoDE & is better than & mRMR & 0.0007\\ \hline
\end{tabular}{}
\end{table}

\subsection{Accuracy}

\begin{figure*}
\center
 \includegraphics[width=0.65\textwidth]{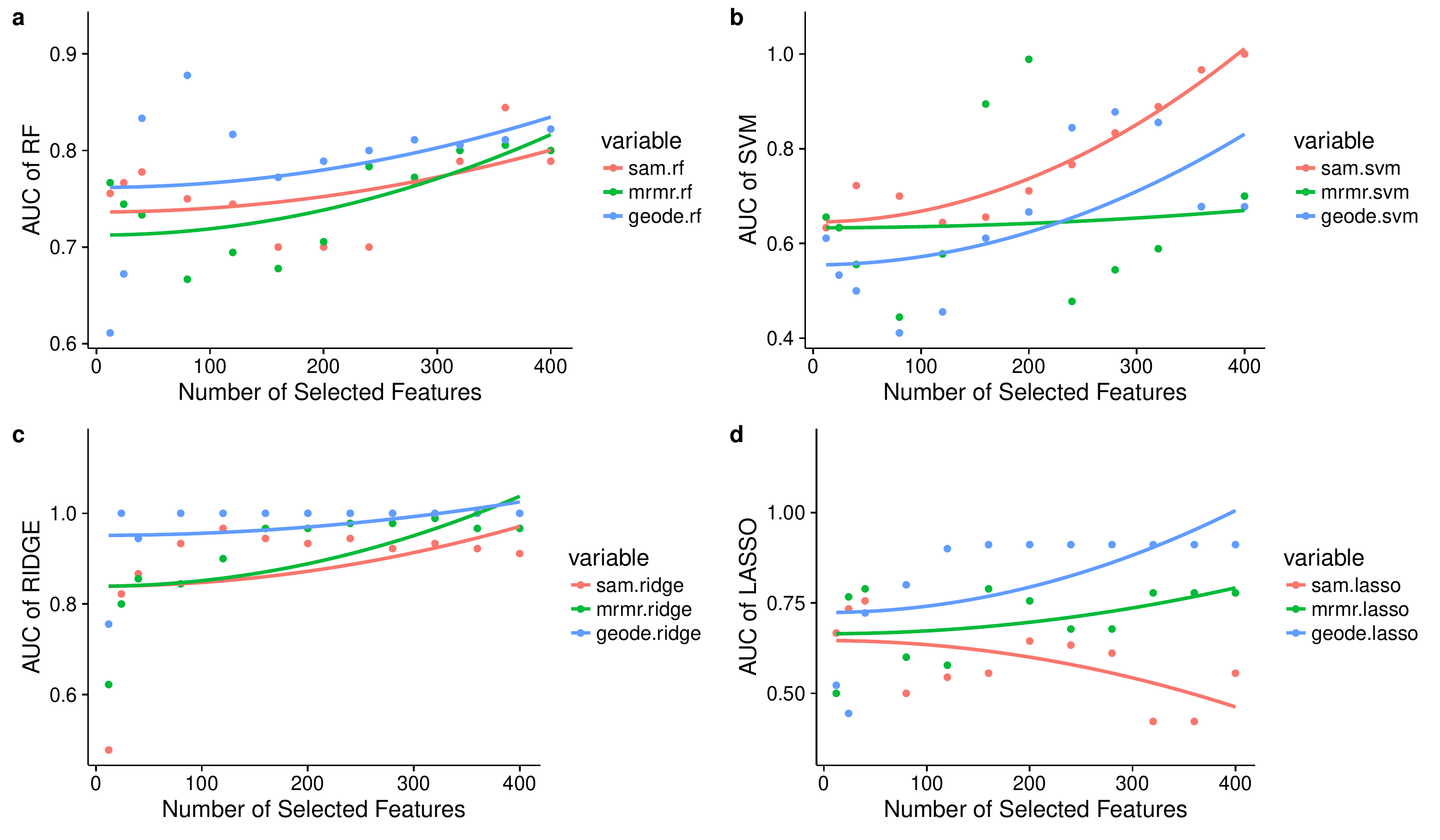}
  \caption{Prediction accuracy on MeHg low dose data.
      (a) using RF
      (b) using SVM
     (c) using RIDGE
     (d) using LASSO. }\label{fig:03}
\end{figure*}

\begin{figure*}
 \includegraphics[width=0.995\textwidth]{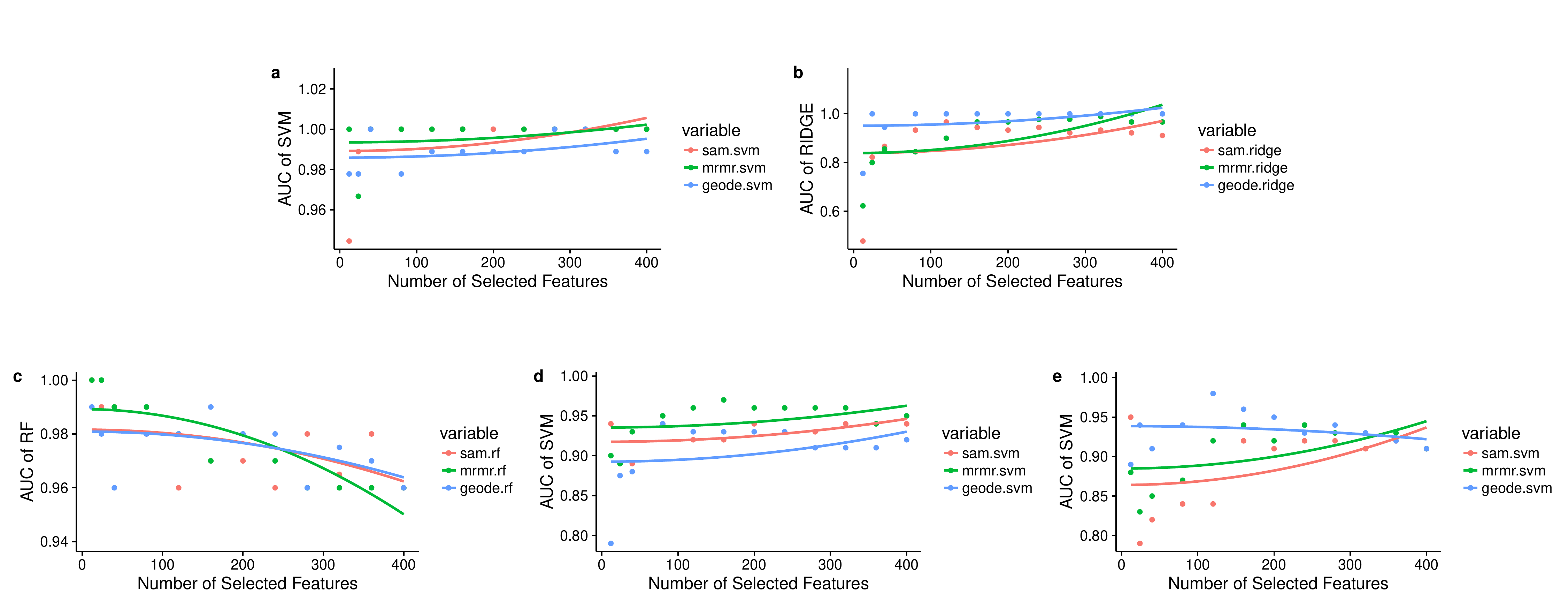}
  \caption{Prediction accuracy.
      (a) in high-dose condition of MeHg
      (b) in low-dose condition of MeHg
     (c) in high-dose condition of PCB 153
     (d) in medium-dose condition of PCB 153
     (e) in low-dose condition of PCB 153. }\label{fig:04}
\end{figure*}

We find that the results of accuracy are not straightforward, since we will get different answers when asking which feature selection method performs the best. In each dose condition, all four classification methods are applied to assess the feature selection methods' ability to improve the prediction accuracy. Across different numbers of selected features, the AUCs of prediction are calculated. Fig.~3\vphantom{\ref{fig:03}} is an example in the condition of low-dose MeHg. It shows that SAM performs the best when the classifier is SVM, but GeoDE turns out to be the best with the other three classifiers. To make it simple, for every experiment (each dose of each toxicant), we select the best classification method for it: a classifier that can give a high prediction accuracy for all three feature selection methods. For example, in low-dose condition of MeHg (Fig.~3\vphantom{\ref{fig:03}}), RIDGE gets the highest prediction accuracy compared with the other three classifiers regardless of the used feature selection method. Then Fig.~4\vphantom{\ref{fig:04}} gives us all results for all conditions. As we can see, different feature selection methods stand out as the best. In low-dose condition of MeHg and PCB 153  (Fig.~4\vphantom{\ref{fig:04}}b, Fig.~4\vphantom{\ref{fig:04}}e), GeoDE performs the best, because it has a higher AUC than the other two in most cases of different numbers of selected features. For the other conditions, in high-dose condition of both MeHg and PCB 153 (Fig.~4\vphantom{\ref{fig:04}}a, Fig.~4\vphantom{\ref{fig:04}}c), and medium-dose condition of PCB 153 (Fig.~4\vphantom{\ref{fig:04}}d), mRMR stands out, especially with a low number of selected features.\par

\begin{table}[!tbp]
\centering
\caption{Tukey HSD Test on prediction accuracy}\label{Tab:02} 
{\begin{tabular}{@{}ccccccc} \hline
Toxicant & Dose condition & Classifier & \multicolumn{3}{c}{Comparison} & p-adj\\ \hline
MeHg & high & RIDGE & mRMR & is better than & GeoDE & 0.0107\\
MeHg & high & RIDGE & mRMR & is better than & SAM & 0.0344\\
MeHg & high & LASSO & mRMR & is better than & GeoDE & 0.0002\\
MeHg & high & RIDGE & SAM & is better than & GeoDE & 0.0003\\
MeHg & low & LASSO & GeoDE & is better than & SAM & 0.0004\\
PCB 153 & high & LASSO & mRMR & is better than & GeoDE & 0.0003\\
PCB 153 & high & LASSO & SAM & is better than & GeoDE & 0.0006\\
PCB 153 & medium & SVM & mRMR & is better than & GeoDE & 0.0077\\
PCB 153 & medium & LASSO & SAM & is better than & GeoDE & 0.0009\\
PCB 153 & medium & LASSO & mRMR & is better than & GeoDE & 0.0009\\
PCB 153 & low & RF & GeoDE & is better than & mRMR & 0.0002\\
PCB 153 & low & RF & GeoDE & is better than & SAM & 0.0082\\
PCB 153 & low & SVM & GeoDE & is better than & SAM & 0.0183\\ \hline
\end{tabular}}{}
\end{table}

Another phenomenon we can see from Fig.~4\vphantom{\ref{fig:04}} is that based on gene expression data and our analysis, MeHg appears to influence cods more than PCB 153 does, since it is easier for classifiers to distinguish between control group and treated group with a small number of features (higher prediction accuracy), and the performance is also more stable.\par

According to the result of Tukey HSD Test on prediction accuracy (Table~\ref{Tab:02}), in different dose conditions and with different classifiers, different feature selection methods will stand out. However, generally speaking, in high-dose condition, mRMR seems to outperform the other two feature selection methods, and in low-dose condition, GeoDE outperforms the other two. \par

\section{Discussion and conclusion}

In this article, we have presented a framework to choose the most suitable biomarker discovery method for a specific dataset by comparing the potential candidates from two aspects: stability, reflecting whether the selected feature subset is robust to changes in the training data, and resulting prediction accuracy. \par

On the aspect of stability to find a reliable feature subset, our results show that GeoDE is more stable than SAM and mRMR in two ways: its stability varies little across different numbers of selected features for all conditions, and the absolute values of stability are always the highest for all numbers of selected features in low-dose condition. \par

On the aspect of feature selection methods' ability to improve a classifier's prediction accuracy, in different dose conditions, different feature selection methods show up as the best. mRMR performs well in high-dose condition, but in low-dose condition, GeoDE outperforms the other two. \par 

To conclude this case study, the choice of the most suitable biomarker discovery method quite depends on the dataset under study. If the experiments are conducted in high dose, then mRMR is the best choice, since it gives the highest prediction accuracy and its stability is comparable with the other two. If it's in low dose, then GeoDE is definitely the best choice, considering its excellent performance both in stability and prediction accuracy. \par

The framework of the comparative analysis is not limited to only this case study, but can be applied to any other similar study. 

%
%

\section*{Acknowledgements}
We would like to thank the colleagues in Jonassen Group for helpful discussions and Computational Biology Unit at University of Bergen, where the work was carried out. We also would like to thank the Centre for Digital Life Norway (DLN) and the dCod 1.0 project to which the work is related. \vspace*{-12pt}

\section*{Funding}

The dCod 1.0 project is funded under the Digital Life Norway initiative of the BIOTEK 2021 program of the Research Council of Norway (project no. 248840).\vspace*{-12pt}

%
%

\end{document}